# Explainable Artificial Intelligence for Smart City Application: A Secure and Trusted Platform

M. Humayn Kabir[1], Khondokar Fida Hasan[2]*, Mohammad Kamrul Hasan[3], Keyvan Ansari[4]

[1]School of Electrical and Electronic Engineering, Islamic University (IU), Kushtia, Bangladesh
[2]School of Computer Science, Queensland University of Technology (QUT), Brisbane, Australia
[3]School of Computer Science, University of Information, Technology and Science (UITS), Dhaka, Bangladesh
[4]School of Science, Technology, and Engineering, University of the Sunshine Coast (USC)
*Corresponding Author, k.fidahasan@yahoo.com

**Abstract.** Artificial Intelligence (AI) is one of the disruptive technologies that is shaping the future. It has growing applications for data-driven decisions in major smart city solutions, including transportation, education, healthcare, public governance, and power systems. At the same time, it is gaining popularity in protecting critical cyber infrastructure from cyber threats, attacks, damages, or unauthorized access. However, one of the significant issues of those traditional AI technologies (e.g., deep learning) is that the rapid progress in complexity and sophistication propelled and turned out to be uninterpretable black boxes. On many occasions, it is very challenging to understand the decision and bias to control and trust systems' unexpected or seemingly unpredictable outputs. It is acknowledged that the loss of control over interpretability of decision-making becomes a critical issue for many data-driven automated applications. But how may it affect the system's security and trustworthiness? This chapter conducts a comprehensive study of machine learning applications in cybersecurity to indicate the need for explainability to address this question. While doing that, this chapter first discusses the black-box problems of AI technologies for Cybersecurity applications in smart city-based solutions. Later, considering the new technological paradigm, Explainable Artificial Intelligence (XAI), this chapter discusses the transition from black-box to white-box. This chapter also discusses the transition requirements concerning the interpretability, transparency, understandability, and Explainability of AI-based technologies in applying different autonomous systems in smart cities. Finally, it has presented some commercial XAI platforms that offer explainability over traditional AI technologies before presenting future challenges and opportunities.

**Keywords**. Explainable AI, Machine Learning, Deep Learning, Cyber Security, Privacy, Transparency.

## 1 Introduction

In the urbanization process worldwide, the concept of smart city gains popularity that promises to offer technology-based services. It adds merit to the traditional existed process; for example, it assists in reducing environmental footprint, improves transportation, increases digital equity, introduces new flows of revenues, improves public utilities and infrastructure, to name a few. Along with the rapid evolution of intelligent technologies and hyper-connectivity, massive amounts of data are likely to be flown through different networks requiring special attention during implementation. The dependency on data and networks makes smart cities prone to cyber-attacks more vulnerable than ever.

The rapid growth in the worldwide urban population and the increasing connectivity of this demographic make the cyber-security of smart cities more critical. This concept will become visible adopting AI with systems and devices. Most of the devices are connected with the network and some of them are utilizing the cloud perform. As a result, there is a high probability of scope to affect with cyber-attack. The explosive device is no longer considered a weapon nowadays. The malicious software named malware can shut down, interrupt, or control the smart space systems.

The scope and extent of cyberspace are increasing day by day. So it's not a human-scale problem to analyze and improve the security issues. We can able to analyze millions of events and identify malware exploiting zero-day vulnerabilities. The risk behavior leading to a phishing attack or malicious code can be easily identified with the help of AI and Machine Learning (ML) technology. The system can learn over time and utilize the past to identify new types of threats using ML. In traditional ML approaches, a data-specific model is developed from learning data. This type of model is applicable for serving particular tasks in a given environment. Due to the rapid progress in the complexity and sophistication of AI-powered systems, it becomes difficult to understand the complex working mechanisms of AI systems in general conditions. Sometimes it is very challenging when AI-based systems compute outputs that are unexpected or seemingly unpredictable. For some ML technology, the autonomy complexity and ambiguity increase, which makes the adaptation and development process interrupted. This especially holds for opaque decision-making systems, which are considered complex black-box models. To overcome these flaws, we need to transfer the classic black box models to white-box models, ensuring ML technology's interpretability, transparency, understandability, and explainability.

Explainable Artificial Intelligence (XAI) is one of the emerging branches of AI that is gaining popularity. It refers to transparent, interpretable, and understandable methods that can produce accurate and explainable models so that humans can understand the offered results [1]. The goal of XAI is to provide a way to explain the cause behind the produced outcome. Arguably, a new generation of AI approaches is emerging using XAI. XAI fundamentally establishes a relationship between the model features and outcome. XAI proposes a set of ML techniques that introduce explainable models, ensuring a high level of learning feedbacks. It maintains security privacy and trust for AI partners. Unlike the classical machine learning approaches, two different elements: explanatory module and explanation interface module, are adopted in XAI.

There is a trade-off between explainability and accuracy. Traditional models can provide better accuracy and efficiency but it has lack explainability. Introducing the XAI can offer the highest level of accuracy, efficiency and as well as explainability. Nowadays, several companies present platforms for Explainable Artificial Intelligence, including IBM, Google, Darwin AI, Flowcast, Imandra, Kybdi, and Factmata, are pioneering. According to a PwC survey reported in 2017, over 70% of the directors in the corporate world keep on the faith of utilizing the benefits of AI with their business. It also claims that AI will enhance the GDP up to $15 trillion by 2030 [2]. Thus, AI technology must become accountable and trustable are the growing demand, where XAI could be the proper choice for the AI community.

At first, this chapter discusses the black-box problems of AI technologies for Cybersecurity applications in smart city-based solutions. Later, this chapter discusses the transition and requirements from black-box to white-box to understand and ensure interpretability, transparency, understandability, and explainability of AI-based technologies in applying autonomous systems in smart cities. It presents some mention of commercial XAI platforms available to offer with traditional AI technologies before presenting XAI challenges and opportunities in the smart city domain.

## 2 Smart Cities: the Better Livings for Dwellers

The idea of smart city originally indicates an enhancement of quality in urban life through technology-based intelligent solutions. Nowadays, about half of the world's population lives in urban areas, and the rate of migration towards the urban areas is increasing significantly. Therefore, the dependency, efficiency, and availability of vital city services and infrastructures such as transport, energy, health care, water, and waste management are growing issues. Smart city aims to deal with all those challenges, including the capacity of city transportation, energy consumption efficiency, maintaining a green environment, improving city dwellers' economic and living standards, are a few from the list. The concept of smart cities is acquiring increasing attention on the agenda of policymakers and gaining momentum. A typical smart city is constructed with several major enabling building blocks, as shown in Fig. 1.

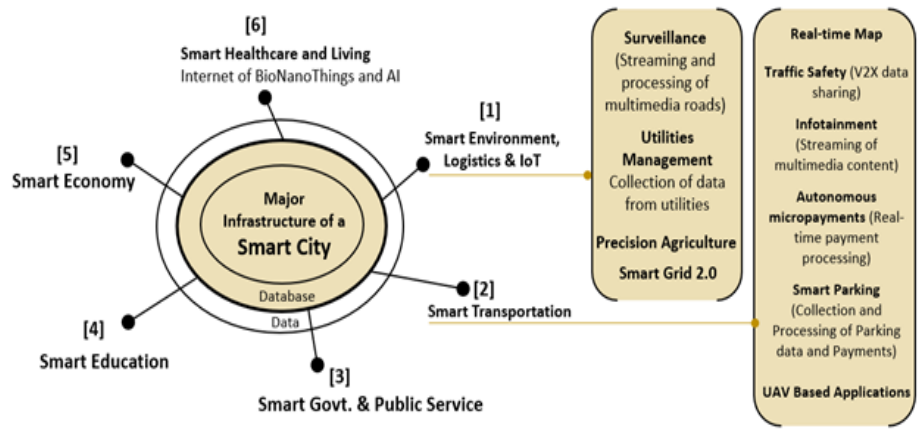

**Fig. 1**. Major enabling blocks of Smart City

Primarily, Information and Communication Technology (ICT) equipped facilitates smart cities to automate and manage city resources and services. It participates in enhancing the city's performance and the welfare of the citizens. Smart city solutions provide intelligent network (Cellular, WiFi, Bluetooth, and Low Power Wide Area (LPWAN)) connectivity and edge processing solutions in cities to connect and improve different infrastructures.

Within the evolved concept of smart city, data can move freely within various connected services. For example, data can generate in the transportation system in response to an emergency to travel to emergency services' networks such as healthcare and government service. Similarly, public utilities (electricity, water, and gas distribution) related data can be helpful to city service operations such as taxation. Such data flows require in development of meaningful processes.

The smart city concept is now growing hand to hand with the rise of Artificial Intelligence (AI). The fundamental role of AI is to enable automated systems to increase the urban system's efficiency and productivity, reduce cost and resource consumption, improve service quality, etc. AI technologies can also serve to overcome some of the burning smart city challenges such as traffic congestion, governance, cybersecurity, and privacy. Table 1 represents some mention of AI applications in different major smart city services.

**Table 1.** Details of Smart City Building Block.

| Services | Description of Service | Notable Application of AI |
|---|---|---|
| Smart Environment, | Smart Environment, logistics, and IoT aim to provide real-time monitoring of the environment and logistics services, including | • Automated statistical information generating. |
| Logistics, and IoT | infrastructure and dispersed asset monitoring. For example, different application types like water and air pollution monitoring and forest fire detection, remote and real-time monitoring of transported goods, a list of few that connect assets from anywhere to any place to make our lives safe and more manageable [3-5]. | • Predicting building energy consumption.<br>• Business Logic generating.<br>• Environment Prediction. |
| Smart Transportation | Smart transportation aims to monitor and manage transportation networks and systems to improve safety and efficiency. It is a combination of multi-functional hyper-connected intelligent transportation systems and advanced digital intelligent transportation information management. For instance, some applications monitor the vehicle load, trip scheduling, passenger entertainment, reservation, toll and ticketing systems, a list of applications alerts for overspeeding, harsh cornering, and acceleration provide a safer road [6]. | • Signal Control and Dynamic route guidance<br>• Traffic Demand Modeling<br>• Identification of drivers based on their activities<br>• Driver's Emotion Detection<br>• Surrounding Object detection |
| Smart. Govt. and Public Service | Smart Govt. is a concept for the public sector that provides collection, connection, and analysis of big data generated and processed in real-time. It uses technologies to build e-government includes a Service-oriented business platform and web applications. The applications of smart government are G2C transactions, civic engagement platforms, online voting systems to stop public gatherings [7]. | • Public to interact with government through the automated virtual assistant<br>• Automatic Disaster detection and management<br>• Postdisaster reconnaissance and identification of affected areas |
| Smart Education | Smart education is an interactive learning model with cloud-based capabilities. It is designed to enhance student engagement with teachers so that teachers can adapt to students' interests and skills. It includes data-centered intelligent education facilities for the learners. It aims to provide digital literacy, effective communication, innovative thinking to develop high-impact projects. It concludes some applications like video clavier, electronic text-books.[8] | • Automated personalized learning.<br>• Differentiated and individualized learning<br>• AI-enabled voice assistant to interact with educational tools. |
| Smart Economy | Smart economy concept combines economic features in sustainable features and eco-economic approaches like economic progress, economic prosperity, sustainable jobs. It aims to improve people's lives and adapt them to social, economic, and environmental policies. It interconnects local and global markets. It provides resource-based development to improve urban economic productivity. Smart economy can | • Monitoring economic growth and ensuring a sustainable employment rate.<br>• Increase Resource efficiency and elevated competitiveness through automated sharing services, |

| | be applied to industry, tourism, maritime, mobility, payment, and banking [4]. | customer-tailored solutions. |
|---|---|---|
| Smart Healthcare and living | Smart healthcare is a service system related to health that uses technology such as wearable Sensor networks, Body Area networks, IoT to access information dynamically and connect healthcare people, materials, and institutions. It helps in the development of telemedicine and creates personalized medical services universally. Smart healthcare is broadly used to assist medical diagnosis and treatment, health management, disease prevention, risk monitoring and management, and provide the patient's result via an automated ledger and communication services [9]. | • Assist physicians with differential diagnosis of diseases.<br>• Prediction model to check up diseases by analyzing patient's electronic health records,<br>• Providing pre-primary care information and remote monitoring to minimize the life risk |

## 3  Security Challenges in Smart City Cyber Domain

In the smart city domain, a wide variety of sensors from extended networks work together to enforce the desired service. Therefore, smart city infrastructure is expected to consist of safe, secure, and reliable interconnected devices such as sensors and actuators to gather, process, and communicate data to enable reliable and trusted city service. However, such network heterogeneity is a great concern for cyber security in the smart city realization always. As on many occasions, small low-end devices with limited capacity can participate in service-enabled communication. Due to processing and storage capacity limitations, such devices may not have robust security features such as authentication mechanisms or cryptographic ability. This leads to a possible attack surface for applications in smart city services.

Therefore, security breaches can be very real in a smart city cyber domain. An intruder can enter the system through cyber-physical system vulnerabilities. A relationship among cyber issues that leads to a cyber threat is shown in Fig. 2.

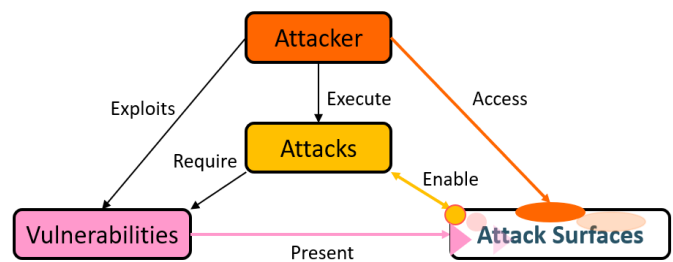

**Fig. 2.** Attack Surface: A relation among cyber security issues [10, 11].

Any system vulnerability can lead to an attack surface. Since in smart cities, both high-end and low-end devices in terms of their capability connect, any Attackers can exploit low-end systems to access the service system. Moreover, any mischief can conduct some series of attacking attempts that reveal an open surface to hack the system unauthorizedly, which is shown in Fig. 2.

In the context of smart cities, cybersecurity fundamentally deals with the security of connected and internet-based services since a large part of the assets are connected to the cloud. However, it also extends the idea to protect the computer system's hardware and software system and the data flow between the networks. It makes the system steady and reliable. Cybersecurity is more effective in protecting external threats in real-time monitoring. As smart cities are the conversion of connected cities, there is a huge amount of data exchange between the entities in this domain. Cities are converged into the digital world for a sustainable economy and environment. The inhabitants of the cities generate a huge amount of data to communicate with their soundings. Smart cities have to support these excessive services and resources. Nowadays, about 55% of the world's population lives in urban areas, and it will increase up to 68% by 2050. The 2018 revision of world urbanization prospects produced by the population division of the UN department of economic and social affairs (UN DESA) notes that due to urbanization, about 2.5 billion world's population shifted to urban areas by 2050. Among them, a 90% increase in population in the metropolitan area will occur in Asia and Africa [12].

To face the needs of this large population, cities have to upgrade the traditional network with IoT technology with the adaptation of AI. This new infrastructure will be able to handle supply chains, assets, and resources management efficiently. It is challenging to handle the sensor network, analysis of data, and reasoning algorithms. Some devices and systems that may not be cyber resilient can impose a threat to security and safety. It's quite a cumbersome job to choose a proper mitigating method that can handle threats. The security issues on IoT devices are well known. There are no security requirements/regulations imposed posed on manufacturers. So, they are not aware of the role of security vulnerability which increases system attack surfaces [13]. The role of smart cities plays a positive impact on business, city services, and people. Middle Eastern cities are now leading to provide smart cities services to their dwellers. For example, Doha is preparing to host the 2022 World Cup by adopting smart transportation and other smart services and applications. Assume that if a cyber-attack is imposed during World Cup 2022 in Doha, everything will be shut down because of a lack of sufficient safety measures [11]. Effective collaboration between vendors, device manufacturers, and governments is needed to fix a security regulation among the IoT community. Emerging standards and guidelines must be adopted to ensure system security during design, testing, and installation. Moreover, the service operator must take steps to understand the security issues and mitigation methods before the incidents occur.

## 4 Classic Artificial Intelligence Methods for Cyber Security in Smart City

Fundamentally, smart cities utilized Internet of Things (IoT) technologies to enable service infrastructure with remote monitoring and control capabilities. Such cyber-physical infrastructure gets technical smartness with automation capabilities by adopting Artificial Intelligence (AI). The heterogeneity of the network in smart city implementation is the greatest threat. And cyber threats can be leading to failure and devastation. The scope of cyber threats is very large. Therefore, along with traditional security measures, machine learning-based AI has also been used to enable security to those cyber-physical systems. A study of ABI research reports that 44% of the cybersecurity expenditure will spend on energy, healthcare, public safety, transportation, water, waste treatment, and other infrastructure management purpose in smart city projects [14]. Within this section's limited scope of the presentation, we will focus on some of the potential threat that deals with AI before stating fundamental challenges.

### 4.1 Fraud Detection

Any financial sector in smart city development is one of the indispensable parts proportionate to the urban economic growth. And this sector is probably the most targeted sector for cyber attacks naturally. Fraud in the financial sector is a common cyber-attack related to financial and reputation problems. Such attacks can be caused by data leakage and credit losses by unauthorized persons. There are several ML approaches present to detect fraud. Authors in [15] presented an SVM classifier and multi-node optimized-based fraud detection model. It used the fraudulent feature to detect the source of fraud involved in e-commerce accurately. This AI-based solution also offers to prevent further fraud. Authors in [16] proposed a Convolution Neural Network (CNN) based fraud detection model, detecting 91% of online fraud activities utilizing online transaction data. They have used the transaction data of a commercial bank. Those are a few of many examples of AI-based approaches to control measures on cyber fraud.

### 4.2 Intrusion Detection

Many research works are noticeable on intrusion detection mechanisms using Artificial Neural Networks, such as Deep Learning (DL) methods. W.H. Lin et al. [17] proposed a cyber intrusion detection system based on the LeNet-5 model. They have utilized over 10,000 training samples reaching an accuracy of 99.65%. Y. Dong et al. [18] presented the Convolution Neural Network (CNN) based AE-Alexnet model utilizing the automatic encoder in real-time cyber intrusion detection system. Their model performs well in intrusion detection utilized the common KDD99 dataset achieve an accuracy of 94.32%. R. U. Khan et al.[19] improved the CNN model for intrusion detection and achieved impressive accuracy of 99.23%, outperforming the DBN model. In this space, there are many successful efforts of utilizing ML to tackle cyber issues[20].

### 4.3 Spam Detection

We are suffering from endless threats of spam which are unwanted emails received from various sources frequently to the users. Bosaeed, et al. proposed a system that can detect short message service spam on mobile devices. They have used Naïve Bayes (NB), Naïve Bayes Multinomial (NBM) and Support Vector Machine (SVM) classifiers. For performance analysis, they have compared the outcome of all the three classifiers, and SVM performed the best among the three classifiers [21]. Meanwhile, deep learning-based spam detection gains popularity. Chetty et al. [22] proposed a neural network-based spam detection model. The experimental results show that this model's performance decreases as the number of datasets decreases. Sharmin et al. [23] proposed another spam detection system using the Multi-Layer Perceptron (MLP), SVM, and CNN. They claim that the CNN model is suitable for detecting spam images with an accuracy of 99.02%. According to the indication of future work, they have concluded that Reinforce Neural Network (RNN) and Long Short Term Memory (LSTM) prove worthwhile in spam detection.

### 4.4 Malware Detection

As smart cities infrastructure is based on interconnecting networks, it is evident that malware is a significant threat that can access the network to compromise the system. Generally, malware is malicious software used to collect users' personal data and digital belongings illegally. The types and threats, complexity, and harmful nature of malware have been devastating day by day. Ransomware, crypto miners draw attention as widespread malware. Yuan et al. presented a malware classifier using Markov Images and Deep Convolution Neural Network (CNN). They have utilized two popular datasets of Microsoft and Drebin. The experimental result shows that the accuracy for Microsoft data is 99.26%, and for Drebin, datasets are 97.36% [24]. Vinayakumar et al.[25]

proposed a DNN based ransomware classifier that used malware datasets collected from tweets from Twitter. They have classified the ransomware malware into 25 classes.

### 4.5 Traffic Analysis and Identification

Mobile devices are generating massive traffic in the IoT domain while using HTTP/HTTPS protocols, one of the prime agents of smart city realization. It is always challenging to identify and analyze any such traffic using traditional methods such as employing hostnames. It is also hard to find the signatures where devices frequently update the software versions. To tackle the traffic issue, Chen, Yu et al. [26] presented a supervised model based on CNN to tackle and identify mobile applications' traffic flow using HTTP sample data packets. This framework gets an average accuracy of 98%. Wang et al. [27] presented a classifier with an intelligent Virtualized Network Functions (VNF) selection model using Deep Neural Network (DNN) and Multi-Grained Cascade Forest. This model was successfully used to schedule the cloud network for vehicular communication to identify the transmission priority of different data packets. Lotfollahi et al. [28] proposed one-dimensional CNN for cyber traffic classifications tasks. They developed a framework based on CNN and Stacked Auto Encoder (SAE). The performance result shows a recall of 0.98 and 0.94 for recognition and traffic classification, respectively.

Overall, the detection of cyber incidences (e.g., attacks and threats) deploying traditional machine learning tools such as deep learning have been widespread in the area of cybersecurity over the last decade. Initially, it is started with conducting code analysis and deploying agents to perform signature-based detection, moving towards more self-learning and automated decompilation and extensive code analysis. It is also enabling the automated analysis and prediction of network monitoring. However, any entire AI-based automatic threat detection is becoming difficult to realize as the complexity and sophistication of the detection system create an additional problem.

Additionally, for strategic and technical reasons, AI-based tools may mislead and cannot detect actual threats. On all those occasions, it may requirer filtering out all known legit network activities to identify the threat using manual analysis. Since conventional AI-based system becomes a higher level of sophistication, it makes them a black box that drastically reduces the amount of information required to conduct manual analysis. This is the major challenge for conventional AI-based threat detection mechanisms. It remains opaque, and it is impossible to understand their back-end function and the process of inference. Therefore, explainability is required that can significantly enhance the detection capabilities with control and transparency.

## 5  Transition to Explainable Artificial Intelligence (XAI)

Machine learning predictive models are treated as black boxes which can be automatically trained without worrying about the domain in which they are used. This opaqueness raises many risks that are difficult to foresee during the model building process. Such as the model's declining performance due to the data drift, poor performance on the out-of-domain problems, or unfairly discriminatory behavior learned on historical data. The growing list of examples where black boxes fail spectacularly has increased interest in XAI methods. Such methods allow X-ray black-boxes models for more detailed analysis on the local or global level.

Adding interpretability to machine learning algorithms is necessary for Transportation, Smart Grid, Healthcare, Industrial applications. The general points of causality, robustness, reliability, trust, and fairness are valid in all domains. Interpretability is primarily a key to identifying causal relationships and increasing the reliability and robustness of scientific discoveries made with the help of ML algorithms. In all application domains, establishing the trust in and fairness of machine learning systems matters most in low-risk environments. In contrast, robustness and reliability are critical to high-risk environments where machines take over decisions with far-reaching consequences.

A general Machine Learning workflow is shown in Fig. 3. Firstly, data is preprocessed to train a model, and after that, the Learning function is built up using the Learning process. Then inputs are utilized to get the learned function which is used to predict the output. New inputs can be fed into the model after learning the function and the machine will return the desired prediction.

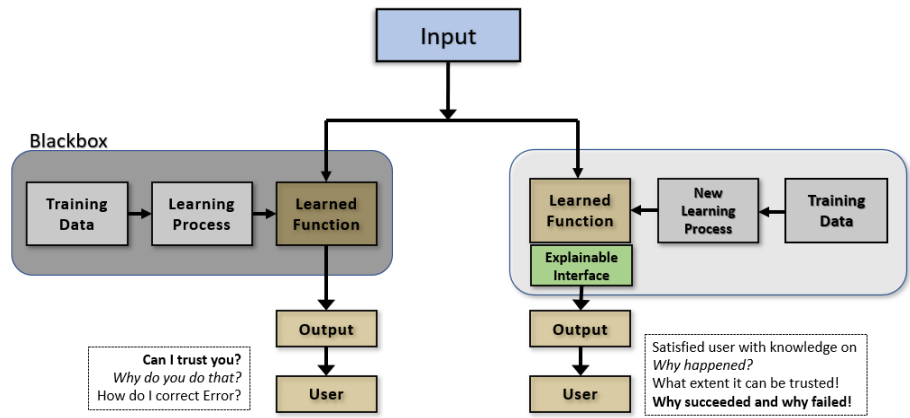

**Fig. 3.** A comparative view of Traditional Machine Learning Workflow (Left) and Explainable Machine Learning Workflow (Right).

The critical thing to note in traditional ML workflow is that its prediction is without informing the reason. It can confuse the user. Users can be misled and have to believe in the machine for making the correct selection. In contrast to the traditional approach, XAI follows an advanced procedure. This procedure provides us the decision with a proper explanation about the background processes. Here the user gets additional information with output which explains the reason for prediction. An extra layer is added, which can explain the model outcome, as shown in Fig. 3. The additional layer helps to ensure impartiality in decision-making because of having an explainability module. If any biases are detected in the datasets, then they can be corrected in this layer. Moreover, explainability discovers the predictions' facts, which is beneficial for finding out weaknesses in outcomes. The methods that can be used to explain the behavior and predictions of trained ML models include, Partial Dependence Plots (PDP), Accumulated Local Effects (ALE), Permutation Feature Importance (PMI), Leave-One-Covariate out (LOCO), and Local Interpretable Model-agnostic Explanations (LIME).

The essence of explainability in AI tools in some major smart city enabling networks is discussed in the following sections.

### 5.1 Transportation System

According to the US Department of Transportation, Intelligent Transportation Systems (ITS) combines different technologies used to monitor, evaluate, and manage transportation systems to enhance efficiency, safety and cost reduction. Smart transportation becomes a reality. Several cities in the world now implement this technology. Cities like New York have adopted smart transportation to build a smart cities. They have also made a testbed for connected vehicles in Wyoming, which is a corridor for the transportation of goods. Introducing smart transportation has improved the efficiency of the supply chain and does not require the driver for the long drive.

Among the several services in smart transportation systems, we will focus on smart vehicles, Intelligent Public Transport (IPT) systems, and smart city traffic management.

Smart vehicles can communicate and exchange data with other vehicles (V2V), pedestrians (V2P), or a generic network or infrastructure (V2N and V2I, respectively) [29]. The integration of IoT devices and 5G communication technologies help to make smart transportation services visible to the city inhabitants. IoT technology, together with sensors networks and embedded controllers, can be used to make any cyber-physical system manage and control remotely.

Smart self-driving cars are taking over decisions in the real world previously taken by humans and can involve severe and sometimes irreversible consequences using machine learning technology. Advanced Driver Assistance Systems (ADASs) technology provides lane-keeping and braking assistants to obtain fully autonomous driverless vehicles [30]. ML can also be an efficient tool for managing the vehicle, e.g., fuel prediction and assessing cybersecurity. IPT operators are also implementing several cybersecurity measures. However, measures are very diverse as there are neither widely accepted cybersecurity standards that align with the needs of IPT nor widely used good practices [31, 32].

In self-driving vehicles, while deploying ML technology, we should be concerned about the algorithm's complexity and be aware of the fool classifier [33]. Time complexity is an important entity in designing the safety-critical context of mobility as it mostly uses embedded technology. Moreover, a false positive result of a classifier may not be acceptable in the case of autonomous driving, such as a stop sign if the classifier treated it as a speed limit [12]. In this case, explainable AI could be a solution that can help to take a decision describing the facts behind it. Explainability could be employed as a way to disclose models' confidential data. At the same time, it also helps to ensure the reliability and robustness of autonomous driving and thus makes it safe.

A. Rosenfeld et al. [34] presented a taxonomy of explainability considering fundamental questions about Why, Who, What, When, and How of explainability. They have utilized this taxonomy in human-agent systems. M. Scalas et al. explore [29] the role of interpretable machine learning in the ecosystem of smart vehicles. They have figured out the terms of explanations that help to design secure vehicles. In [34] the authors define explainability from an engineering view by explaining the problem of AI models. They have identified the problem of AI models in different autonomous vehicles domain for object detection, control, and perception. They proposed to introduce explainability features to the existing AI models. E. Soares et al. [35] proposed a new explainable self-organizing architecture and a new density-based feature selection method for the autonomous vehicle. They have focused on a self-organizing neuro-fuzzy approach to learn interpretable models automatically.

**5.2 Healthcare**

Proper networking between city health services using ICT infrastructure can be helpful to view the health status of community residents. Mobile devices can collect personal health data and can support private healthcare efficiently. Users can utilize a smartphone to collect health-related information using various Health apps. Moreover, users can get information about medical insurance packages, nutrition, fitness, drugs, and doctors browsing websites. The patient's observation and monitoring of disease symptoms and behavioral patterns (activity, sleep disorder, mobility pattern, phone conversation) become easy using the flavor of IoT. The mobile healthcare system can detect the changes of different patient symptoms and take the safety measure accordingly both online and remotely [36]. Mobile devices can play an important role, from individual monitoring to community-level monitoring. A medical practitioner can keep important record notes on diagnosis for a patient. Severe and chronically ill people can utilize social media to share their illness, remedy experiences, and support the community. Some users may not prefer physical support; in that case, online support using mobile devices and IoT is helpful.

People who are suffering from long or short-term chronic diseases can be tracked using mobile ICT. Users can also keep a record of their drug-taking schedule and can set alarms as a reminder. Awareness can be patronized among the inhabitants as prevention. Online medical technologies are getting popular among city dwellers using mobile devices and apps. Moreover, the integration of ambient sensors into smart city networks makes it possible to provide services to vulnerable and non-tech-savvy patients. City designers and researchers urge for integrating smart ICT and Healthcare. The advancement in pervasive computing and machine learning accelerates the integration of smart healthcare in smart city technology. Now vast amounts of personal health data are shared; as a result, privacy issues arise. Although healthcare system utilizing Artificial Intelligence (AI) has shown, it's competency in providing cost-effective and human-centric application. The performance of medical diagnosis, image-video processing, resource tracking can be improved by introducing the power of AI. Therefore, the convergence of AI technologies to healthcare is a demand of current application in the smart cities domain.

Recent advancement in AI technology demonstrates remarkable performance in implementing AI and machine learning in healthcare. Healthcare requires a high level of accountability and transparency. Security and privacy are two main issues that are the barriers to accommodating technology in healthcare. The current ML solutions are mostly dark in nature based on the black-box model, which doesn't become explainable. It is against the moral responsibility of physicians to use the black-box medical decision-making system [37]. This system can not provide the explanation behind the reasoning. In the case of consulting an expert AI system, the physician may ask which factors are considered of reasoning. The traditional AI system, which is opaque in nature, can not provide the background behind the reasoning outcome to the physician. In order to provide the answer of the physician, the expert system has to build the knowledge of explanation that satisfies the physician. Thus explanation provides reliability, improves transparency in the decision-making system.

In [9], authors have presented suggestions for explainable medical AI systems. They explored the idea of explanation in view of human-centric orientation. S. Meacham et al. [38] have designed and developed a web system to predict and explain patient readmittance using machines. They have utilized a logistic regression-based ML algorithm with an explainability function and a web interface corresponding users' interfaces.

### 5.3 Smart Grid

The increasing population in cities causes a decrease in fossil energy sources. So renewable energy sources are the best solution for supporting the energy demand. About 75-80% of the energy of the world is consumed in cities. This energy consumption generates about 80% of the world's greenhouse gas emissions [7]. Smart Grid (SG) is the key item of energy infrastructure for sustainable smart cities. It aims to reform fossil energy resources into renewable energy resources. SG plays a critical role in reducing environmental pollution due to the use of non-renewable energy resources.
Moreover, it can provide lower energy costs and reliable energy service. The traditional non-smart grid lacks online monitoring and controlling facilities which is a flaw for a real-time solution. The smart city energy system should have facilities to connect each energy system and grid for real-time remote control and monitoring. This ensures efficient energy generation and consumption. A prediction shows that the 15.8% smart energy market will invest around US$248.36 billion for producing smart energy systems [39]. In a review paper [40], the smart plug technology is illustrated, potentially contributing to green energy, SG/microgrids, and smart cities. In [41], the authors presented current state-of-the-art techniques used to predict advanced intelligent loads in SG.

Smart meters provide monitoring of the exact and real-time electricity usage, offering bill information adopting Advanced Metering Infrastructure (AMI). It also helps to insist on carbon emissions and assists in making intelligent decisions. Smart grids support Electric Vehicles (EVs) to detect and accept the produced/stored energy from the users' side. Smart grid offers smart load and building automation, leading to increased energy efficiency, safety, and comfort. The smart city project of Malaga, Spain, launched in 2009, aims to be the largest energy-smart city [42]. It utilized smart meters for remote management to improve energy efficiency. A light-emitting diode (LED) network was also built up for street lighting. Another project of future smart cities started in 2015 at the EU named Europe's future smart cities. Eindhoven, Manchester, and Stavanger, in the Netherlands, UK, and Norway, respectively, are equipped with LED networks and serve as testbeds. The plan is to imply the findings to the other three cities Leipzig, Sabadell, and Prague, in Germany, Spain, and Czech Republic, respectively [38]. Several technologies related to energy, carbon emission, and air quality are imposed on this area, and the performance is closely observed. The main focus of this project is to update the SGs by integrating real-time monitoring, automation, and self-controlling facilities.

AI approaches have been applied to various smart grid applications, such as demand response, predictive maintenance, and load forecasting. However, traditional AI techniques are considered a ''black-box'' due to their lack of explainability and transparency. AI-based methods are utilized to predict the energy consumption aims to save energy [43]. Consider the case; a heating device could be turned off if the device fails to operate on the low power source. However, the heating device can also be turned off to save energy when the desired ambient temperature is reached or when the predicted energy consumption is reached above the threshold. This scenario can be modeled using XAI model only, which can explain the real fact why the heating device is turned off. In traditional black-box AI techniques, it is not possible to know the fact behind the reasoning. XAI could be an interesting research idea because it connects with the expert to the finding of the AI system by explaining the cause behind the reasoning.

In [44], the authors proposed an approach to introduce XAI in ontologies and semantic technology to provide autonomous car decision-making and energy efficiency solutions. M. Kuzlu et al. [11] utilize three XAI tools (LIME, SHAP, and ELI5) [33] to produce high-level explanations about the outcome of the system in smart grid applications.

## 6      Commercial XAI Platforms

XAI technology is emerging. Several commercial platforms are formed to support this increasing demand. In this section, we will look at some of the popular XAI platforms in brief.

**IBM.** IBM is one of the pioneers of developing XAI platforms which has opened an advanced cloud-based AI tool using XAI technology. It aims to expose the causes behind the present state of art AI recommendations. They conducted an extensive survey that reported that over 60% of its executives feel uncomfortable with the AI solution's traditional black-box approach. In other ways, such a survey inspires them to propel the explainability in their AI solutions [45].

IBM ensures fairness, robustness, explainability, accountability, and value alignment in AI applications. IBM aimed to utilize these virtues in all respects to the whole lifecycle of their AI application. IBM integrated XAI frameworks and toolsets into the IBM Cloud Paks, which supports the data platform. They use AIx 360 toolkit, which includes a set of algorithms that cover several domains of explanations. This tool is made for algorithmic research for different practices of domains. As a result, their AI technology supports businesses in a controlled, safe and measurable way. The AI practitioner can build, run, and manage AI models and can optimize decisions anywhere on IBM Cloud named Paks® using IBM Watson® Studio for Data [46]. The application area of IBM using XAI is Healthcare, Financial Services, and Education. IBM Cloud Paks can increase the model accuracy by 15% to 30%.

**Google.** Another XAI-enabled AI platform has been launched by tech giant Google, which explains the performance of several of Google's features. The feature attributions are generated using AutoML Tables and AI Platform for model predictions in XAI solution platform. Whereas the model behavior can be investigated visually using the What-If Tool. The AI practitioner, such as data scientists, can explore to improve the usable datasets or models. The model performance is also debugged using AI Explanations in AutoML Tables, AI platform predictions, and AI platform notebooks provided by Google.

Moreover, the model behavior can be investigated using the What-If Tool minutely. The ultimate target is to build users' trust and improve the transparency and fairness of machine learning models through human-interpretable explanations. We can predict deploying an AutoML table or AI platform, providing a real-time score [47]. The ground truth of predictions input can be sampled using continuous evaluation capability from the trained ML models. Mainly the XAI-enabled platform designed by google explains the image recognition and code sampling features of google. Arguably, Google has a pervasive influence on smart city realization from an intelligent car to smart home development.

**Darwin AI.** Drawin AI, an XAI service provider company, enables enterprises to structure AI on which they can stand. Darwin AI is active in building a process known as Generative synthesis. It aims to make the explainability real, granting the makers to realize the interior functions of deep neural networks. Within this framework, Darwin AI employs an artificial neural network-based deep learning method. It offers an Explainability toolkit and network diagnostics features. Moreover, it also provides some other network automation for connectivity applications. It focuses on detecting defects and adaptive factory automation, one of the significant smart infrastructure sectors.

**Flowcast.** Flowcast is an API-based solution that targets bringing out the black box models by fusing different company systems. Flowcast clarifies the relationships between input and output values of different patterns to establish explainability. It utilizes its own AI technologies to develop credit assessment models with the explainable module. Such an enhancement comes up with the potential to transform credit regulations completely into the new ability of explainability. Among a selection of ML algorithms used by Flowcast, the boosted trees algorithm is an opaque variant; they had to find a way to explain system assessments clearly. Thus, explainability uses SHAP along with NLP to provide plaintext phases presenting outputs in layman's terms. Therefore, help financial institutions adhere to regulatory compliance by reducing risk and unlocking credit for all.

**Imandra.** Imandra is an automated reasoning technology that is a digitization and AI for capital market APIs. Imandra democratizes automated reasoning to enable algorithms to be explainable, trustworthy. They present the Reasoning as a Service (RAAS) feature, which fetches these methods nearer to those except a specialized background in these areas. Initially, Imandra started with the financial sector and has worked with major investment players. It has focused on service designing, testing, and conducting AI-based audits of its complex trading systems, which has a great future in smart city applications.

**Kyndi.** Kyndi is a pioneering text analytics company that utilizes an explainable natural language processing platform. It optimizes human cognitive performance by transforming business processes through auditable AI systems. Kyndi provides the auditable trail of reasoning if the stakeholders of organizations seek an explanation of the output. Kyndi uses a well-established programming language called Prolog with RPA tools to build bots that automate fact, inference, and concept generation in almost any vertical. Its expanded model aids in encoding the semantics of documents, repositories, and domains in query-able knowledge graphs. Kyndi aims to utilize its tools for smart city application in association with different stakeholders, including federal government sectors, such as Defense, Intelligence, finance, healthcare, IT, and infrastructure.

**Factmata.** Fake news is created maliciously to misinform people by generating more traffic to sites, maybe publishing scandalous headlines or politically biased content focusing on chaos. Content Moderation performed by humans is time-consuming, repetitive, and open to errors. Factmata is an AI solution that makes AI power tools detect problematic content from inter. Factmata is fighting against online fake news with XAI techniques to justify segregation [48].

## 7    Opportunities and Challenges of XAI

The Explainability model could significantly increase the detection competency, and the prediction result could be trustworthy. Many of the application domains are handling personal and legal issues data. Hence aware should be taken to find the cause and issues behind the outcome of reasoning provided by the trained model. The model involved in explainability should not generate a large set of negative results. Much concentration should be focused on explaining the trained model in prediction and classification applications. In open-source intelligence, explainability could provide a means for detecting attacks, impacting mitigating their influence [49].

Nowadays, ML models used in cyber security do not have an Explanations module. It is very important to find the cause behind reasoning because system administrators can demand more information other than informing only the binary decision. A realistic threat model can be utilized to know about the strength of mitigation methods in the cyber-security domain. There are several XAI methods available nowadays. The good part is that these XAI methods can expose the mode of processing to produce the decision of trained ML models to some extent. Explainability would also have a significant impact on Adversarial Machine Learning as well.

However, several research challenges exist to develop methods for explainability from the technical, legal, and practical points of view. Technical challenges not only deal with XAI system development but also evaluation and interpretation. It is because the XAI system must include an explicit representation of the component parts to support the appropriate and systematic interpretation of working functions. In such cases, a set of semantics representations can help construct an explanation that a human receiver understands. It is, however, may still tricky for a human to interpret the functioning of the XAI methods. Therefore, XAI systems should aim to deal with the role of the learning method. It should have a set of open procedures to reason the inference process. Overall, explanations of the AI model are needed for the proper design and functioning of the correct decision-making system. Although, they are not the ultimate goal of designing AI system. XAI applies knowledge and expertise to design a sensible and transparent AI system.

The application of XAI methods in any specific domain, on the other hand, can face the question of legitimacy. This can happen when handling data involving legal issues, decisions, life, and finance, gathering more intelligence on the people involved. The explainability module in a learning model utilizes to infer data about entities. These allowed can produce significant data, underlying or hidden correlations, and causality is a challenge.

The success of XAI depends on fruitful human-machine interaction. It is considered that both machines' outcome human decision would be consistent irrespective of the ground truths. However, there is high uncertainty in the ground truth and cannot be fully explained. Current XAI methods and techniques predominantly depend on input–relevant parts and do not incorporate the human model's notion [50]. Hence the concept of causability arises, which is a challenge for XAI realization.

## 8    Conclusion

This chapter presented an overall analysis of the prospect of explainable artificial intelligence in the cyber security domain of smart city applications. We conducted our study to portray the opportunity and necessity of XAI using secondary data. Undoubtedly, the advancement of AI is primarily dependent on the flourishment of ML techniques. It will pave the path to handle the high-frequency, real-time application for the smart city. It is, however, difficult to see how such methods work in underlining. This is very real to combat any cyber security issues. If any AI-based cyber security component does not have forensic value but remains as a black box, it certainly has less credibility than all. Therefore, XAI has a great role in ensuring smart city deployment's security, privacy, and trustworthiness.

## References

[1]    D. V. Carvalho, E. M. Pereira, and J. S. J. E. Cardoso, "Machine learning interpretability: A survey on methods and metrics," vol. 8, no. 8, p. 832, 2019.

[2]    D. A. S. R. a. G. Verweij. "Sizing the prize What's the real value of AI for your business and how can you capitalise?" https://www.pwc.com/gx/en/issues/analytics/assets/pwc-ai-analysis-sizing-the-prize-report.pdf (accessed.

[3]    M. H. Kabir, M. R. Hoque, and S.-H. J. I. J. o. S. H. Yang, "Development of a smart home context-aware application: a machine learning based approach," vol. 9, no. 1, pp. 217-226, 2015.

[4]    J.-L. Briaud et al., "Realtime monitoring of bridge scour using remote monitoring technology," Texas Transportation Institute, 2011.

[5]    K. F. Hasan, Y. Feng, and Y.-C. Tian, "GNSS time synchronization in vehicular ad-hoc networks: Benefits and feasibility," IEEE Transactions on Intelligent Transportation Systems, vol. 19, no. 12, pp. 3915-3924, 2018.

[6]     C. B. Y. Zhang, Y. Zhang, J. Zhang and J. Xu, "Automatic Mobile Application Traffic Identification by Convolutional Neural Networks," in IEEE Trustcom/BigDataSE/ISPA, 2016, pp. 301-307, doi: 10.1109/TrustCom.2016.0077.

[7]     T. Nam and T. A. Pardo, "Smart city as urban innovation: Focusing on management, policy, and context," in Proceedings of the 5th international conference on theory and practice of electronic governance, 2011, pp. 185-194.

[8]    "Smart City Malaga." https://malagasmart.malaga.eu/en/sustainable-and-safe-habitat/energy/smartcity-malaga/ (accessed 2021/09/07.

[9]    Y. Xie, G. Gao, and X. A. J. a. p. a. Chen, "Outlining the design space of explainable intelligent systems for medical diagnosis," 2019.

[10]    K. Fida Hasan, A. Overall, K. Ansari, G. Ramachandran, and R. Jurdak, "Security, Privacy and Trust: Cognitive Internet of Vehicles," arXiv e-prints, p. arXiv: 2104.12878, 2021.

[11]    M. Kuzlu, U. Cali, V. Sharma, and Ö. J. I. A. Güler, "Gaining insight into solar photovoltaic power generation forecasting utilizing explainable artificial intelligence tools," vol. 8, pp. 187814-187823, 2020.

[12]    U. D. o. E. a. S. A. U. DESA). "The Revision of the World Urbanization Prospects, Population Division of the United Nations." https://www.un.org/development/desa/en/news/population/2018-revision-of-world-urbanization-prospects.html (accessed.

[13] I. Butun, P. Österberg, H. J. I. C. S. Song, and Tutorials, "Security of the Internet of Things: Vulnerabilities, attacks, and countermeasures," vol. 22, no. 1, pp. 616-644, 2019.

[14] J. Wilson and N. J. I. T. o. M. C. Patwari, "Radio tomographic imaging with wireless networks," vol. 9, no. 5, pp. 621-632, 2010.

[15] A. S. Sadiq, H. Faris, A.-Z. Ala'M, S. Mirjalili, and K. Z. Ghafoor, "Fraud detection model based on multi-verse features extraction approach for smart city applications," in Smart cities cybersecurity and privacy: Elsevier, 2019, pp. 241-251.

[16] Z. Zhang, X. Zhou, X. Zhang, L. Wang, P. J. S. Wang, and C. Networks, "A model based on convolutional neural network for online transaction fraud detection," vol. 2018, 2018.

[17] W.-H. Lin, H.-C. Lin, P. Wang, B.-H. Wu, and J.-Y. Tsai, "Using convolutional neural networks to network intrusion detection for cyber threats," in 2018 IEEE International Conference on Applied System Invention (ICASI), 2018: IEEE, pp. 1107-1110.

[18] Y. Dong, R. Wang, and J. He, "Real-time network intrusion detection system based on deep learning," in 2019 IEEE 10th International Conference on Software Engineering and Service Science (ICSESS), 2019: IEEE, pp. 1-4.

[19] R. U. Khan, X. Zhang, M. Alazab, and R. Kumar, "An improved convolutional neural network model for intrusion detection in networks," in 2019 Cybersecurity and cyberforensics conference (CCC), 2019: IEEE, pp. 74-77.

[20] M. Z. Alom and T. M. Taha, "Network intrusion detection for cyber security on neuromorphic computing system," in 2017 International Joint Conference on Neural Networks (IJCNN), 2017: IEEE, pp. 3830-3837.

[21] S. Bosaeed, I. Katib, and R. Mehmood, "A Fog-Augmented Machine Learning based SMS Spam Detection and Classification System," in 2020 Fifth International Conference on Fog and Mobile Edge Computing (FMEC), 2020: IEEE, pp. 325-330.

[22] G. Chetty, H. Bui, and M. White, "Deep learning based spam detection system," in 2019 International Conference on Machine Learning and Data Engineering (iCMLDE), 2019: IEEE, pp. 91-96.

[23] T. Sharmin, F. Di Troia, K. Potika, and M. J. I. S. J. A. G. P. Stamp, "Convolutional neural networks for image spam detection," vol. 29, no. 3, pp. 103-117, 2020.

[24] S. Yang et al., "Scalable digital neuromorphic architecture for large-scale biophysically meaningful neural network with multi-compartment neurons," %J IEEE transactions on neural networks learning systems, vol. 31, no. 1, pp. 148-162, 2019.

[25] R. Vinayakumar, M. Alazab, K. Soman, P. Poornachandran, A. Al-Nemrat, and S. J. I. A. Venkatraman, "Deep learning approach for intelligent intrusion detection system," vol. 7, pp. 41525-41550, 2019.

[26] Z. Chen, B. Yu, Y. Zhang, J. Zhang, and J. Xu, "Automatic mobile application traffic identification by convolutional neural networks," in 2016 IEEE Trustcom/BigDataSE/ISPA, 2016: IEEE, pp. 301-307.

[27] J. Wang, B. He, J. Wang, and T. J. I. T. o. V. T. Li, "Intelligent VNFs selection based on traffic identification in vehicular cloud networks," vol. 68, no. 5, pp. 4140-4147, 2018.

[28] M. Lotfollahi, M. J. Siavoshani, R. S. H. Zade, and M. Saberian, "Deep packet: A novel approach for encrypted traffic classification using deep learning," %J Soft Computing, vol. 24, no. 3, pp. 1999-2012, 2020.

[29] M. Scalas and G. Giacinto, "On the Role of Explainable Machine Learning for Secure Smart Vehicles," in 2020 AEIT International Conference of Electrical and Electronic Technologies for Automotive (AEIT AUTOMOTIVE), 2020: IEEE, pp. 1-6.

[30] N. Dasanayaka, K. F. Hasan, C. Wang, and Y. Feng, "Enhancing Vulnerable Road User Safety: A Survey of Existing Practices and Consideration for Using Mobile Devices for V2X Connections," arXiv preprint arXiv:2010.15502, 2020.

[31] K. F. Hasan, C. Wang, Y. Feng, and Y.-C. Tian, "Time synchronization in vehicular ad-hoc networks: A survey on theory and practice," Vehicular Communications, vol. 14, pp. 39-51, 2018.

[32] K. F. Hasan, T. Kaur, M. M. Hasan, and Y. Feng, "Cognitive internet of vehicles: motivation, layered architecture and security issues," in 2019 International Conference on Sustainable Technologies for Industry 4.0 (STI), 2019: IEEE, pp. 1-6.

[33] T. Gu, B. Dolan-Gavitt, and S. J. a. p. a. Garg, "Badnets: Identifying vulnerabilities in the machine learning model supply chain," 2017.

[34] F. Hussain, R. Hussain, and E. J. a. p. a. Hossain, "Explainable Artificial Intelligence (XAI): An Engineering Perspective," 2021.

[35] E. Soares, P. Angelov, D. Filev, B. Costa, M. Castro, and S. Nageshrao, "Explainable density-based approach for self-driving actions classification," in 2019 18th IEEE

[35] International Conference On Machine Learning And Applications (ICMLA), 2019: IEEE, pp. 469-474.

[36] D. J. Cook, G. Duncan, G. Sprint, and R. L. J. P. o. t. I. Fritz, "Using smart city technology to make healthcare smarter," vol. 106, no. 4, pp. 708-722, 2018.

[37] A. J. London, "Artificial intelligence and black-box medical decisions: accuracy versus explainability," J Hastings Center Report, vol. 49, no. 1, pp. 15-21, 2019.

[38] S. Meacham, G. Isaac, D. Nauck, and B. Virginas, "Towards explainable AI: design and development for explanation of machine learning predictions for a patient readmittance medical application," in Intelligent Computing-Proceedings of the Computing Conference, 2019: Springer, pp. 939-955.

[39] A. Gaviano, K. Weber, and C. J. E. P. Dirmeier, "Challenges and integration of PV and wind energy facilities from a smart grid point of view," vol. 25, pp. 118-125, 2012.

[40] N. K. Suryadevara and G. R. Biswal, "Smart plugs: Paradigms and applications in the smart city-and-smart grid," %J Energies, vol. 12, no. 10, p. 1957, 2019.

[41] S. N. Fallah, R. C. Deo, M. Shojafar, M. Conti, and S. J. E. Shamshirband, "Computational intelligence approaches for energy load forecasting in smart energy management grids: state of the art, future challenges, and research directions," vol. 11, no. 3, p. 596, 2018.

[42] J. L. Y. Wang, Y. Chen, M. Gruteser, J. Yang, and H. Liu, "E-eyes: Device-free location-oriented activity identification using fine-grained WiFi signatures," in Proc. Annu. Int. Conf. Mob. Comput. Networking, MOBICOM, 2014, pp. 617-628, doi: 10.1145/2639108.2639143.

[43] N. Petrović and Đ. J. S. J. o. E. E. Kocić, "Data-driven framework for energy-efficient smart cities," vol. 17, no. 1, pp. 41-63, 2020.

[44] N. Petrović and M. Tošić, "EXPLAINABLE ARTIFICIAL INTELLIGENCE AND REASONING IN SMART CITIES OBJAŠNJIVA VEŠTAČKA INTELIGENCIJA I REZONOVANJE U PAMETNIM GRADOVIMA."

[45] M. De Sanctis, E. Cianca, S. Di Domenico, D. Provenziani, G. Bianchi, and M. Ruggieri, "Wibecam: Device free human activity recognition through wifi beacon-enabled camera," in Proceedings of the 2nd workshop on Workshop on Physical Analytics, 2015, pp. 7-12.

[46] IBM. "Explainable AI. Available." https://www.ibm.com/se-en/watson/explainable-ai (accessed 04-08-2021.

[47] G. Cloud. "Explainable AI." https://cloud.google.com/explainable-ai (accessed.

[48] Factmata. "Helping Organizations Understand Online Content." https://www.factmata.com/ (accessed.

[49] S. M. Devine and N. D. Bastian, "Intelligent systems design for malware classification under adversarial conditions," arXiv preprint arXiv:1907.03149, 2019.

[50] A. Holzinger, G. Langs, H. Denk, K. Zatloukal, and H. Müller, "Causability and explainability of artificial intelligence in medicine," Wiley Interdisciplinary Reviews: Data Mining and Knowledge Discovery, vol. 9, no. 4, p. e1312, 2019.